%% file: iclr2025_conference.tex
\documentclass{article} 
\usepackage{float}
\usepackage{graphicx}
\usepackage{epstopdf}
\usepackage{algorithm}
\usepackage{multirow}
\usepackage{multicol}
\usepackage{algpseudocode} 
\usepackage{makecell} 
\usepackage{iclr2025_conference,times}
\usepackage{subcaption} 

\input{math_commands.tex}

\usepackage{hyperref}
\usepackage{url}

\title{Linear Preference Optimization: Decoupled Gradient Control via Absolute Regularization}

\author{Rui Wang\thanks{These authors contributed equally to this work.}, Qianguo Sun$^{*}$, Chao Song, Yu Li\thanks{Corresponding author.} \\
International Digital Economy Academy \\
\texttt{\{wangrui,sunqianguo,songchao,liyu\}@idea.edu.cn} \\
\And
Junlong Wu, Tianrong Chen, Zhiyun Zeng \\
Emdoor Collaborative Laboratory \\
\texttt{\{tianrong.chen, zhiyun.zeng, junlong.wu\}@emdoor.com} \\
}

\begin{document}

\maketitle
\begin{abstract}
DPO (Direct Preference Optimization) has become a widely used offline preference optimization algorithm due to its simplicity and training stability. However, DPO is prone to overfitting and collapse. To address these challenges, we propose Linear Preference Optimization (LPO), a novel alignment framework featuring three key innovations. First, we introduce gradient decoupling by replacing the log-sigmoid function with an absolute difference loss, thereby isolating the optimization dynamics. Second, we improve stability through an offset constraint combined with a positive regularization term to preserve the chosen response quality. Third, we implement controllable rejection suppression using gradient separation with straightforward estimation and a tunable coefficient that linearly regulates the descent of the rejection probability. Through extensive experiments, we demonstrate that LPO consistently improves performance on various tasks, including general text, math, text-to-speech (TTS), and automatic speech recognition (ASR) tasks. These results establish LPO as a robust and tunable paradigm for preference alignment. Source code and model checkpoints are publicly available at \href{https://github.com/IDEA-Emdoor-Lab/LPO}{https://github.com/IDEA-Emdoor-Lab/LPO}.
\end{abstract}

\section{Introduction}
The alignment of large language models (LLMs) with human preferences has become a critical step in developing capable and safe AI assistants. Reinforcement Learning from Human Feedback (RLHF) \cite{ouyang2022rlhf}, particularly through proximal policy optimization (PPO) \cite{schulman2017ppo}, has established the dominant paradigm for this alignment. Although effective, PPO suffers from significant complexity, requiring multiple models (reward model, reference policy, active policy) and intricate online sampling and optimization processes, leading to high computational costs and implementation instability. To address these limitations, Direct Preference Optimization (DPO) \cite{rafailov2023direct} emerged as a simpler and more stable alternative. DPO reframes preference learning as a supervised loss function directly applied to the policy network, bypassing the need for explicit reward modeling or online RL.

Despite its elegance and widespread adoption, DPO exhibits several critical shortcomings. First, the inherent coupling within the log-sigmoid function forces the optimization of log-probability of the chosen \( \log \pi_\theta(y_w | x) \) and the rejected response \( \log \pi_\theta(y_l | x) \) to be interdependent. This often manifests itself as an undesirable and significant decrease in the logarithmic probability of the responses chosen during training, which can degrade their inherent quality even as the preference objective improves. Second, DPO is highly sensitive to the quality and noise level within preference datasets. Suboptimal or ambiguous preference pairs can lead to overfitting and subpar performance. Third, DPO lacks explicit mechanisms to control the magnitude of the gap between the logarithmic probabilities of the chosen and rejected responses \( \log \frac{\pi_\theta(y_w | x)}{\pi_{\text{ref}}(y_w | x)} - \log \frac{\pi_\theta(y_l | x)}{\pi_{\text{ref}}(y_l | x)} \), which can lead to overoptimization and reduced generalization.

To overcome these fundamental limitations of DPO, we propose Linear Preference Optimization (LPO), a novel preference alignment algorithm built upon three key innovations:
\begin{itemize}
    \item \textbf{Gradient Decoupling via Absolute Regulation}: We replace the 'log-sigmoid' function with the absolute difference function. This crucial modification decouples the gradients flowing back to the chosen and rejected log-probabilities, enabling more independent and targeted optimization of each term.
    \item \textbf{Stability Enhancement via Offset and Positive Constraint}: Inspired by Offset-DPO (ODPO) \cite{amini2024odpo} and Identity Preference Optimization (IPO) \cite{azar2023ipo}, we introduce an offset \( \mu \) to explicitly constrain the gap \( \log \frac{\pi_\theta(y_w | x)}{\pi_{\text{ref}}(y_w | x)} - \log \frac{\pi_\theta(y_l | x)}{\pi_{\text{ref}}(y_l | x)} \), preventing it from growing excessively large and improving generalization. Simultaneously, inspired by DPOP \cite{pal2024dpop}, we incorporate an explicit positive constraint term \( -\log \pi_\theta(y_w | x) \) to counteract the problematic decrease in log-probability of chosen responses observed in standard DPO.
    \item \textbf{Controlled Rejection Suppression via Gradient Separation}: Leveraging the Straight-Through Estimator (STE) technique \cite{esser2021vqgan}, we strategically detach the computational graph (using `tensor.detach()`) to isolate the gradients of the chosen and rejected log-probabilities. This allows us to introduce a control coefficient \( r_2 \) specifically on the gradient path influencing the log-probability of the rejected response. By modulating \( r_2 \), we gain fine-grained control over the rate at which the log-probability of rejected responses is suppressed during optimization.
\end{itemize}

We conducted extensive experiments to validate the effectiveness of LPO. Our results demonstrate:
\begin{itemize}
    \item \textbf{General Capability}: LPO achieves state-of-the-art or highly competitive performance on general instruction-following benchmarks, including MT-Bench \cite{bai2024mtbench} and AlignBench \cite{liu2023alignbench}, confirming its robustness and general applicability.
    \item \textbf{Specialized Task Superiority}: LPO excels in specialized domains. In particular, models fine-tuned with LPO significantly outperform strong baselines such as Qwen2.5-Instruct on mathematical reasoning (GSM8K \cite{cobbe2021gsm8k}). Furthermore, LPO yields substantial performance gains in specialized speech processing tasks, including Automatic Speech Recognition (ASR) and Text-to-Speech (TTS) synthesis within a native speech multimodality framework.
    \item \textbf{Controllability Validation}: Ablation studies confirm the efficacy of the \( r_2 \) coefficient in precisely regulating the suppression rate of rejected response log-probabilities, providing practitioners with a valuable tuning knob for alignment behavior.
\end{itemize}

In summary, our work makes the following key contributions:
\begin{itemize}
    \item We introduce \textbf{Linear Preference Optimization (LPO)}, a novel DPO variant designed to address core limitations: gradient coupling, chosen response degradation, and uncontrolled gap growth.
    \item We propose three core innovations: 1) absolute difference for gradient decoupling, 2) offset constraint and positive regularization for stability, and 3) STE-based gradient separation with a rejection control coefficient \( r_2 \) for tunable optimization dynamics.
    \item We provide comprehensive empirical evidence showcasing LPO's effectiveness across general instruction-following, specialized mathematical reasoning, and speech processing tasks.
    \item We demonstrate and validate the practical utility of the coefficient \(r_2\) for controlling rejection suppression rates.
\end{itemize}

\section{Related Works}
Current large language models (LLMs) demonstrate a strong capability in following human instructions \cite{yang2025qwen3,liu2024deepseek}, showcasing their effectiveness in various applications such as text generation, question answering, and conversational agents. These models leverage extensive training on diverse datasets, allowing them to understand and respond accurately to user inputs. Their performance is further enhanced through techniques like Reinforcement Learning from Human Feedback (RLHF), which aligns model outputs with human preferences \cite{schulman2017proximal}. This continuous refinement of LLMs not only improves their responsiveness but also ensures they adhere to ethical guidelines and user expectations. However, RLHF training necessitates the separate training of a reward model \cite{christiano2017deep, ouyang2022training}, which in turn requires the simultaneous loading of four distinct models. This method can be resource-intensive and may lead to instability during the training process \cite{rafailov2023direct}. To mitigate these issues, Direct Preference Optimization (DPO, \cite{rafailov2023direct}) introduces a novel parameterization method for the reward model. This innovative approach facilitates the derivation of the optimal policy through a closed-form solution, enabling traditional RLHF challenges to be addressed using a straightforward classification loss function.

The DPO training process is prone to overfitting, with both positive and negative sample probabilities tending to decrease \cite{feng2024towards}. To address these issues, several improvements have been proposed, such as DPOP \cite{pal2024smaug} and IPO \cite{azar2024general}. IPO examines the theoretical underpinnings of RLHF and DPO. To address DPO’s tendency to overfit, IPO introduces a pairwise preference loss function known as “Identity Preference Optimization (IPO).” This function is designed to mitigate overfitting on preference data by penalizing preference margins that exceed a specified regularization threshold, thereby enhancing the model’s generalization capabilities. DPOP \cite{pal2024smaug} addresses the issue of declining positive sample probability by incorporating a penalty term for positive samples into its objective function. SimPO \cite{meng2024simpo} utilizes the average log-probability of a sequence as its implicit reward function, which allows for more accurate alignment of the model’s generation behavior. This reward structure not only enhances the effectiveness of the model but also eliminates the need for a reference model. As a result, SimPO significantly boosts computational efficiency and minimizes memory usage, making it a more resource-friendly approach in preference optimization. Our approach modifies DPO by replacing the log-sigmoid function with an absolute function. We also integrate SimPO’s length normalization and decouple the gradient computations for positive and negative samples. This decoupling provides explicit control over the magnitude of gradient descent for negative samples, enabling more targeted optimization and improving the overall performance of the model.

\section{Methods}
\subsection{Limitations of DPO}
RLHF \cite{ouyang2022rlhf} has been demonstrated to significantly enhance the robustness of model responses, safety, and reasoning capabilities of large language models in previous training paradigms. However, the practical application of RLHF presents several challenges. These include the difficulty in training a sufficiently robust reward model to prevent reward hacking, while certain RLHF algorithms, such as Proximal Policy Optimization (PPO) \cite{schulman2017ppo}, suffer from difficult parameter tuning. Consequently, to streamline the alignment training process for large models, Direct Preference Optimization (DPO) \cite{rafailov2024dpo} has been proposed. DPO demonstrates that the RLHF training procedure can be simplified into a maximum likelihood optimization problem without the need to explicitly train a separate reward model:

\begin{equation}
\label{dpo_eq}
\begin{split}
L_{\text{DPO}}(\pi_\theta, \pi_{\text{ref}}) = 
&- E_{(x, y_w, y_l) \sim D} \bigg[ \log \sigma \bigg( \beta \cdot \log \frac{\pi_\theta(y_w | x)}{\pi_{\text{ref}}(y_w | x)} - \beta \cdot \log \frac{\pi_\theta(y_l | x)}{\pi_{\text{ref}}(y_l | x)} \bigg) \bigg]
\end{split}
\end{equation}

In Eq.\ref{dpo_eq}, \(\sigma\) signifies the logistic function; \(D = \{(x^i, y_w^i, y_l^i)\}_{i=1}^N\) represents the dataset, where \(x^i\) represents the prompt, and \(y_w^i\) and \(y_l^i\) denote the chosen response and rejected response for the input prompt \(x\) respectively; The term \(\pi_\theta\) refers to the policy model to be optimized, which is initialized from the Supervised Fine-Tuning (SFT) model, while \(\pi_{\text{ref}}\) denotes the Reference model, also derived from the SFT model.

Based on the gradient analysis \cite{feng2024dpolimits} of DPO, we perform a variable substitution in Eq.\ref{dpo_eq} to facilitate subsequent gradient analysis:
\begin{equation}
\label{dpo_eq_sub}
\begin{split}
L_{DPO}(\pi_\theta, \pi_{ref})=-E_{(x, y_w, y_l) \sim D} \big[ \log \sigma ( \beta x_1 - \beta x_2 ) \big]
\end{split}
\end{equation}

In Eq.\ref{dpo_eq_sub}, \(x_1\) represents \(\log\frac{\pi_{\theta}(y_w|x)}{\pi_{\text{ref}}(y_w|x)}\), and \(x_2\) represents \(\log\frac{\pi_{\theta}(y_l|x)}{\pi_{\text{ref}}(y_l|x)}\).
Then, we take the partial derivatives with respect to \(x_1\) and \(x_2\) respectively, and we can obtain:
\begin{equation}
\label{dpo_eq_derive}
\begin{split}
\left\{
\begin{aligned}
\frac{\partial L_{DPO}(x_1, x_2)}{\partial x_1} = & -\frac{\beta x_2^{\beta}}{x_1(x_1^{\beta} + x_2^{\beta})}  \\
\frac{\partial L_{DPO}(x_1, x_2)}{\partial x_2} = & -\frac{\beta x_2^{\beta - 1}}{x_1^{\beta} + x_2^{\beta}}
\end{aligned}
\right.
\end{split}
\end{equation}

In Eq.\ref{dpo_eq_derive}, \(\frac{\partial L_{DPO}(x_1, x_2)}{\partial x_1}\) is divided by \(\frac{\partial L_{DPO}(x_1, x_2)}{\partial x_2}\), and then we can obtain the following:

\begin{equation}
\label{dpo_eq_divide}
\begin{split}
\left|\frac{\partial L_{DPO}(x_1, x_2)}{\partial x_1} \bigg/ {\frac{\partial L_{DPO}(x_1, x_2)}{\partial x_2}}\right| = \frac{x_2}{x_1}
\end{split}
\end{equation}

Based on the characteristics of the BT model and the DPO training objectives, we can conclude that to maximize the likelihood function of the DPO, the condition \(x_1 - x_2\) must be positive, implying that \(x_1 > x_2\). Consequently, the gradient produced by \(x_1\) (corresponding to the chosen response) will be less than that generated by \(x_2\) (corresponding to the rejected response). Additionally, due to the marginal effect of the logistic function, in the later stages of training, \(x_2\) will become significantly smaller than \(x_1\), leading to the gradient from \(x_2\) dominating the training process. This results in the log-probabilities of the rejected responses being reduced to extremely low values. However, in practice, there is no necessity to decrease the log-probabilities of the rejected responses to such an extent.

Meanwhile, since the loss of DPO essentially increases \(x_1 - x_2\), the following situations may arise in the trends of \(x_1\) and \(x_2\):

\begin{equation}
\label{x1_x2_trend}
\begin{cases} \text{Case 1: } x_1 \uparrow, x_2 \downarrow, \text{the rate of \(x_1\) rising is slightly higher than \(x_2\) descending} \\[2ex] \text{Case 2: } x_1 \downarrow, x_2 \downarrow, \text{the rate of \(x_1\) descending is lower than \(x_2\)} \\[2ex] \text{Case 3: } x_1 \uparrow, x_2 \uparrow, \text{the rate of \(x_1\) rising is higher than \(x_2\)} \end{cases}
\end{equation}

Among the three cases, Case 1 represents the ideal optimization target for DPO, where \(x_1\) slightly increases while \(x_2\) decreases at a reasonable rate. However, in practical DPO training, a common issue arises where both \(x_1\) and \(x_2\) decrease simultaneously. This simultaneous decline can lead to a reduction in model performance, undermining the effectiveness of the training process.

The limitations of DPO can be summarized in two key aspects:

(i) The contribution of the chosen responses to the gradient is consistently less than that of the rejected responses. This imbalance results in the optimization process predominantly focusing on reducing the log-probabilities of the rejected samples. Consequently, the nature of the sigmoid function exacerbates this issue, leading to an excessively large reduction in the log-probabilities of the rejected responses.

(ii) Since the objective of DPO training is fundamentally to increase the difference between \(\log \frac{\pi_{\theta}(y_w|x)}{\pi_{\text{ref}}(y_w|x)}\) and \(\log \frac{\pi_{\theta}(y_l|x)}{\pi_{\text{ref}}(y_l|x)}\), it frequently leads to both values decreasing simultaneously. This simultaneous decline results in a reduction in the model’s performance rather than an improvement, undermining the effectiveness of the training process.

\subsection{Linear Preference Optimization: Decoupling the grad between chosen and reject}

Eq.\ref{dpo_eq} shows that the DPO target function can be represented as \(L_{DPO}(x_1, x_2) = f(x_1, x_2)\). From Eq.\ref{dpo_eq_sub}, the gradients \(\frac{\partial L_{DPO}(x_1, x_2)}{\partial x_1}\) and \(\frac{\partial L_{DPO}(x_1, x_2)}{\partial x_2}\) incorporate nonlinear terms involving both \(x_1\) and \(x_2\). Therefore, we proceed to linearize the mathematical expression of DPO to facilitate further analysis and optimization.

To enhance LPO function, we replace DPO’s log-sigmoid function with the absolute function. We also introduce an offset inspired by IPO \cite{azar2023ipo} and ODPO \cite{amini2024odpo}, incorporate a positive term motivated by DPOP \cite{pal2024dpop}, and apply length normalization to both chosen and rejected log-probabilities similar to SimPO \cite{meng2024simpo}. The resulting LPO function can be expressed as follows:

\begin{equation}
\label{lpo_eq}
\begin{split}
\mathcal{L}_{LPO} = 2\beta \cdot \left| x_1 - x_2 - \frac{1}{2\beta} \right| + \lambda \cdot \max (0, -x_1)
\end{split}
\end{equation}

Here, \(\beta\) and \(\lambda\) are hyperparameters controlling the offset and the magnitude of the positive term, respectively, while \(x_1\) and \(x_2\) represent the length-normalized log-probabilities of the chosen and rejected responses:

\begin{equation}
\label{lpo_x1_x2}
\begin{split}
\begin{cases}
x_1 = \log \frac{\pi_\theta(y_w|x)^{\frac{1}{len_w}}}{\pi_{\text{ref}}(y_w|x)^{\frac{1}{len_w}}} = \frac{1}{len_w} \cdot \log \frac{\pi_\theta(y_w|x)}{\pi_{\text{ref}}(y_w|x)} \\[2ex]
x_2 = \log \frac{\pi_\theta(y_l|x)^{\frac{1}{len_l}}}{\pi_{\text{ref}}(y_l|x)^{\frac{1}{len_l}}} = \frac{1}{len_l} \cdot \log \frac{\pi_\theta(y_l|x)}{\pi_{\text{ref}}(y_l|x)}
\end{cases}
\end{split}
\end{equation}

Where \(len_w\) denotes the length of the chosen response, and \(len_{l}\) denotes the length of the rejected response.

Using the gradient analysis method \cite{feng2024dpolimits}, we can compute the partial derivatives of LPO function with respect to the variables \(x_1\) and \(x_2\). The derivatives can be expressed as follows:

\begin{equation}
\label{lpo_derivation}
\begin{split}
\begin{cases} 
\frac{\partial L_{LPO}(x_1, x_2)}{\partial x_1} = -2\beta \cdot \text{sgn}(x_1-x_2-\frac{1}{2\beta}) + C \\[3ex]
\frac{\partial L_{LPO}(x_1, x_2)}{\partial x_2} =-2\beta \cdot \text{sgn}(x_1-x_2-\frac{1}{2\beta}) 
\end{cases}
\end{split}
\end{equation}

Where \text{sgn}($u$)  is the sign function, which is defined as 1 if \( u > 0 \), -1 if \( u < 0 \), and 0 if \( u = 0 \). The expression for \( C \) is also a constant, as shown in Eq. \ref{lpo_constant}. This constant plays a crucial role in adjusting the gradient based on the value of \( x_1 \), ensuring that the optimization process is influenced appropriately depending on whether the chosen response log-probability is negative or not. These partial derivatives allow us to analyze how changes in the log-probabilities of the chosen and rejected responses affect the overall optimization objective, providing insights into the dynamics of the optimization process.

\begin{equation}
\label{lpo_constant}
\begin{split}
C = \lambda \cdot \begin{cases} 1 & \text{if } x_1 < 0 \\ 0 & \text{if } x_1 \ge 0 \end{cases} \\[3ex]
\end{split}
\end{equation}

We divide \(\frac{\partial L_{L P O}\left(x_1, x_2\right)}{\partial x_1}\) by \(\frac{\partial L_{L P O}\left(x_1, x_2\right)}{\partial x_2}\) to obtain the following expression:

\begin{equation}
\label{lpo_divide}
\begin{split}
\frac{\partial L_{L P O}\left(x_1, x_2\right)}{\partial x_1} \bigg/ \frac{\partial L_{L P O}\left(x_1, x_2\right)}{\partial x_2}=\frac{-2\beta \cdot \text{sgn}(x_1-x_2-\frac{1}{2\beta}) + C}{2\beta \cdot \text{sgn}(x_1-x_2-\frac{1}{2\beta})}
\end{split}
\end{equation}

This simplification reveals that the ratio of \(x_1\) and \(x_2\) becomes a constant, and the relative magnitude of their gradients can be controlled by \(\beta\) and \(\lambda\). Meanwhile, to more effectively control the descent rate of \(x_2\), we utilize the Straight-Through Estimator (STE) as introduced in \cite{esser2021vqgan}. This technique allows us to enhance Eq. \ref{dpo_eq} by enabling gradient propagation through discrete operations while maintaining the ability to optimize continuous variables effectively. As following:

\begin{equation}
\label{lpo_ste_x1}
\begin{split}
\begin{cases}
L_{LPO-ste}^{x_1} = r_1 \cdot 2\beta \left| x_1 - x_2.\text{detach}() - \frac{1}{2\beta} \right| + \lambda \cdot \max(0, -x_1) \\[2ex]
L_{LPO-ste}^{x_2} = r_2 \cdot 2\beta \left| x_1.\text{detach}() - x_2 - \frac{1}{2\beta} \right| + \lambda \cdot \max(0, -x_1.\text{detach}())  
\end{cases}
\end{split}
\end{equation}

By applying the STE, we can isolate the gradients of the chosen and rejected log-probabilities, allowing for separate adjustment of their descent rates. This separation enhances the flexibility of our optimization process, facilitating finer control over the learning dynamics and improving overall model performance.

Ultimately, the expression for LPO-ste can be formulated as follows:

\begin{equation}
\label{lpo_ste_final}
\begin{split}
\begin{cases} 
L_{LPO-ste} = \frac{2}{r_1 + r_2} \cdot \left( r_1 \cdot L_{LPO-ste}^{x_1} + r_2 \cdot L_{LPO-ste}^{x_2} \right) \\[2ex]
x_1 = \frac{1}{len_w} \cdot \log \frac{\pi_\theta(y_w | x)}{\pi_{ref}(y_w | x)} \\[2ex]
x_2 = \frac{1}{len_l} \cdot \log \frac{\pi_\theta(y_l | x)}{\pi_{ref}(y_l | x)} 
\end{cases}
\end{split}   
\end{equation}

In this expression, \( L_{x_1}^{LPO-STE} \) and \( L_{x_2}^{LPO-STE} \) represent the losses corresponding to the chosen and rejected responses, respectively, while \( r_1 \) and \( r_2 \) are coefficients that control the descent rates for these two components. By using the STE, we ensure that gradients are effectively managed during the optimization process, allowing for improved performance in preference alignment tasks.

\begin{figure}[htbp] 
\centering
\includegraphics[scale=0.4]{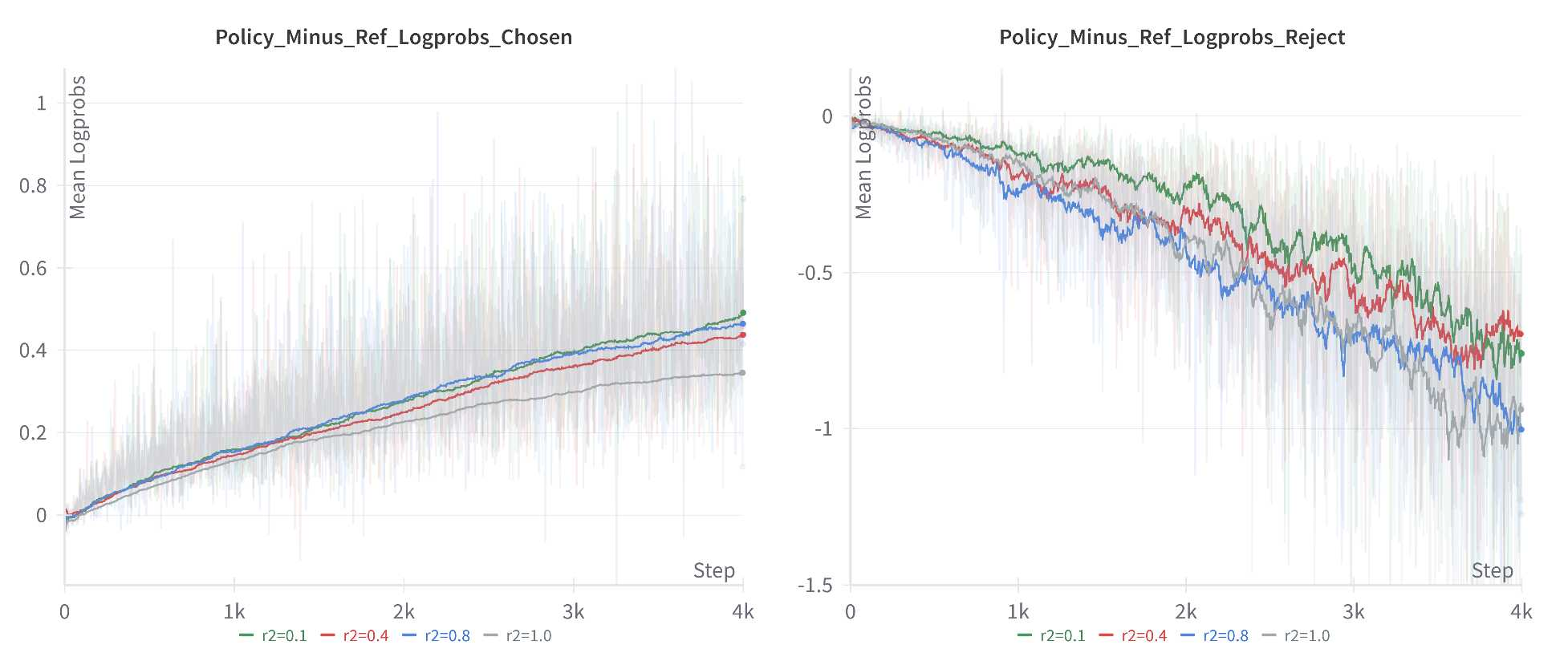}
\caption{The changes of chosen and reject are shown when \(r_2\) takes values of 0.1, 0.4, 0.8, and 1.0. A downward trend in reject can be seen. As \(r_2\) increases, the descent rate grows. Meanwhile, the upward trend of chosen, which corresponds to the above, shows a reduction. This matches the theoretical analysis of the relative gradient changes of LPO.}
\label{lpo_x1_x2_fig}
\end{figure}

In the practical application of LPO-ste, \(r_1\) is typically fixed at 1.0, while \(r_2\) is adjusted within the range \([0.05, 3.0]\). Fig. \ref{lpo_x1_x2_fig} illustrates the descent rate of the rejected responses and the ascent rate of the chosen responses under varying \(r_2\) values. From Fig. \ref{lpo_x1_x2_fig}, it is evident that adjusting the size of \(r_2\) facilitates control over the ascent rate of the chosen responses and the descent rate of the rejected responses. This capability enables us to fine-tune the model’s performance effectively.

\subsection{Preference Pairs Construction}

In the SPIN \cite{chen2024spin}, it is noted that after SFT training, a general model’s output still displays certain discrepancies when compared to the Ground Truth. Iterative DPO \cite{pang2024idpo} suggests that a reward model can be utilized to select samples with the highest and lowest scores for constructing preference pairs.

Consequently, we propose a novel method for constructing preference pairs without relying on a reward model. In this approach, the chosen sample is considered a sufficiently good answer, while the rejected sample is generated using the SFT model’s inference hyperparameters, with both top-p and temperature set to 1.0.

The specific algorithm for this preference pair construction is detailed in Algorithm \ref{alg_lppc}.

This method allows for the effective generation of preference pairs while reducing dependency on additional models, enhancing efficiency in the optimization process.

\begin{algorithm}
\caption{LPO Preference Pair Construction (LPPC)}
\label{alg_lppc}
\begin{algorithmic}[1]
\Require
\State $\mathcal{D} = \left\{x^i, y^i\right\}_{i=1}^N$: where $x^i$ represents the prompt of the preference optimization dataset and $y^i$ represents the corresponding Ground Truth.
\State $\pi_\theta(x)$: the Supervised Fine-tuning (SFT) model.
\Ensure
\State \textbf{Step 1: Construct Chosen}: 
\State \quad $\mathcal{D_{\text{chosen}}} = \left\{x^i, y_{\text{chosen}}^i \equiv y^i\right\}_{i=1}^N$, where $y_{\text{chosen}}^i$ is always equal to the corresponding $y^i$.
\State \textbf{Step 2: Construct Reject}: 
\State \quad $\mathcal{D_{\text{reject}}} = \left\{x^i, y_{\text{reject}}^i \equiv \pi_\theta(x^i | topp=1.0, temp=1.0)\right\}_{i=1}^N$
\State \textbf{Output:}$\mathcal{D}_{\text{LPPC}} = \left\{x^i, y_{\text{chosen}}^i \equiv y^i, y_{\text{reject}}^i \equiv \pi_\theta(x^i)\right\}_{i=1}^N$
\end{algorithmic}
\end{algorithm}

\section{Experiments}

In this section, we first introduce the experimental setup and subsequently present the comparative experimental results of the models. To validate the effectiveness of the proposed algorithm, comprehensive experiments were conducted across three distinct domains: general text tasks, domain-specific mathematical text tasks, text-to-speech (TTS) speech generation tasks, and automatic speech recognition (ASR) tasks.

The experimental framework involved selecting a diverse set of benchmarks to assess the performance of the models effectively. Each domain was meticulously designed to evaluate specific capabilities:
\begin{enumerate}
    \item \textbf{General Text Tasks}: These tasks focused on a variety of language understanding and generation challenges, including writing, summarization, and question answering.

    \item \textbf{Domain-Specific Mathematical Text Tasks}: This segment aimed to test the models on mathematical reasoning and problem-solving capabilities, utilizing datasets specifically curated for this purpose.

    \item \textbf{TTS Speech Generation Tasks}: In this area, models were evaluated on their ability to synthesize natural-sounding speech, focusing on fidelity, stability, and emotional expressiveness.

    \item \textbf{ASR Speech Recognition Tasks}: In this domain, models were assessed on their proficiency in converting spoken language into accurate and robust textual transcriptions, with an emphasis on handling diverse accents, background noise, and varying speech rates.
    
\end{enumerate}

The results will be presented in terms of performance metrics relevant to each task type, showcasing the improvements achieved through the application of the proposed method compared to baseline models.

\subsection{Results on General Tasks} \label{test_on_general_task_lpo}

We selected Qwen2.5 \cite{team2024qwen2} as the base model for our experiments. Infinity-Instruct \cite{li2025infinity} serves as an open-source, high-quality dataset. For SFT phase, we randomly selected 290k examples from the Infinity-Instruct dataset. The model resulting from this SFT process is denoted as qwen2.5-SFT. Detailed configurations and the training procedure for the SFT are provided in the Appendix \ref{sft_configurations}.

During the alignment phase, we validated the algorithm's robustness using both noisy training data and high-quality preference training data.

1) Infinity-Preference: This is an open-source, high-quality preference dataset characterized by subtle distinctions between chosen and rejected responses. It features minimal noise and presents greater learning difficulty, making it an ideal candidate for robust evaluation.

2) Infinity-instruct-1w: For this dataset, we randomly selected 10,000 samples from the remaining Infinity-Instruct data. The responses from the original Infinity-Instruct dataset are designated as the chosen set, while the rejected set comprises data generated by Qwen2.5-SFT, utilizing temperature and top-p parameters set to 1.0. It is important to note that this constructed training dataset is of lower quality compared to Infinity-Preference.

Previous research has shown that GPT-4 evaluations closely align with human assessments and are significantly more cost-effective \cite{zheng2023judging}. Consequently, we selected two GPT-4-based evaluation benchmarks: MT-bench \cite{bai2024mtbench} and Alignbench \cite{liu2023alignbench}.

MT-bench encompasses a variety of categories, including writing, STEM, role-play, reasoning, mathematics, humanities, information extraction, and coding, while also assessing multi-turn dialogue performance. For this benchmark, we employ GPT-4 as the judge model. In a similar vein, Alignbench includes evaluation categories such as mathematical computation, role-playing, logical reasoning, and text writing.

For comprehensive tasks, we conducted detailed experiments on both MT-bench and Alignbench. In both benchmarks, we employ GPT-4 as the evaluation judge model.

We employ vanilla DPO \cite{rafailov2023direct} as our baseline comparison algorithm. For the DPO experiments, we adhere to the official setup, configuring the \( \beta \) hyperparameter to 0.1. The experimental setup for LPO is detailed in Appendix Table \ref{tab:param_lpo_general}.

\begin{table}[h]
\centering
\small
\setlength{\tabcolsep}{4pt}
\caption{Details of LPO performance on MT-Bench trained on infinity-preference dataset}
\label{tab:performance_mt_bench_on_perference}
\begin{tabular}{ccccccccccc}
\hline
model & Turn & writing & stem & roleplay & reasoning & math & humanities & extraction & coding & avg \\
\hline
\multirow{2}{*}{SFT} & 1 & 9.1 & 8.7 & 8.2 & 6.6 & 8.5 & 9.2 & 8.8 & 5.5 & \multirow{2}{*}{7.65} \\
 & 2 & 6.7 & 7.3 & 7.7 & 5.3 & 5.6 & 9.4 & 8.9 & 7.0 & \\
\hline
\multirow{2}{*}{DPO} & 1 & \textbf{9.2} & \textbf{9.3} & \textbf{8.8} & 6.4 & \textbf{9.2} & 9.2 & 8.6 & 7.0 & \multirow{2}{*}{\textbf{8.20}} \\
 & 2 & 8.3 & 7.7 & 8.4 & 5.5 & 6.2 & \textbf{9.8} & \textbf{10.0} & 7.6 & \\
\hline
\multirow{2}{*}{LPO} & 1 & 9.1 & 8.9 & 8.5 & \textbf{8.1} & 8.8 & 8.9 & 8.1 & \textbf{7.9} & \multirow{2}{*}{8.16} \\
 & 2 & 7.9 & 7.6 & 8.0 & 6.3 & 6.8 & 9.5 & 8.8 & 7.4 & \\
\hline
\end{tabular}
\end{table}

\begin{table}[h]
\centering
\small
\setlength{\tabcolsep}{3pt}
\caption{Details of LPO performance on AlignBench trained on infinity-perference dataset}
\label{tab:performance_alignbench_on_perference}
\begin{tabular}{c|ccc}
\hline
Task & SFT & DPO & LPO \\ \hline
Professional Skill & 6.62 & \textbf{7.12} & 6.59 \\
Chinese Comprehension & 5.82 & \textbf{6.25} & 6.13 \\
Basic Task & \textbf{6.45} & 6.22 & 6.35 \\
Math Computation & 6.45 & \textbf{6.65} & 6.49 \\
Text Writing & 5.65 & 5.21 & \textbf{6.16} \\
Comprehensive Q\&A & 6.23 & 7.23 & \textbf{7.26} \\
Roleplay & 6.55 & 6.61 & 6.92 \\
Logical Reasoning & 5.66 & 5.46 & \textbf{5.89} \\
Chinese Reasoning & 6.06 & 6.05 & \textbf{6.14} \\
Chinese Language & 6.22 & \textbf{6.61} & 6.57 \\ \hline
\textbf{Overall Score} & 6.14 & 6.34 & \textbf{6.36} \\ \hline
\end{tabular}
\end{table}

\begin{table}[htbp]
\centering
\small
\setlength{\tabcolsep}{4pt}
\caption{Details of LPO performance on MT-Bench trained on infinity-instruct-1w dataset}
\label{tab:performance_mt_bench_on_instruct}
\begin{tabular}{ccccccccccc}
\hline
model & Turn & writing & stem & roleplay & reasoning & math & humanities & extraction & coding & avg \\
\hline
\multirow{2}{*}{SFT} & 1 & \textbf{9.1} & 8.7 & 8.2 & 6.6 & \textbf{8.5} & 9.2 & 8.8 & 5.5 & \multirow{2}{*}{7.65} \\
 & 2 & 6.7 & 7.3 & 7.7 & 5.3 & 5.6 & \textbf{9.4} & 8.9 & \textbf{7.0} & \\
\hline
\multirow{2}{*}{DPO} & 1 & 8.9 & 8.7 & \textbf{8.7} & 7.3 & \textbf{8.5} & 8.5 & 9.7 & 6.1 & \multirow{2}{*}{7.63} \\
 & 2 & 7.4 & 7.1 & 7.7 & 5.9 & 5.5 & 9.1 & 7.3 & 5.7 & \\
\hline
\multirow{2}{*}{LPO} & 1 & 9.0 & \textbf{9.1} & 8.4 & \textbf{8.2} & 8.4 & 9.0 & 8.8 & 5.4 & \multirow{2}{*}{\textbf{8.02}} \\
 & 2 & 8.3 & 7.5 & 7.9 & 7.5 & 5.7 & 9.3 & \textbf{9.9} & 6.0 & \\
\hline
\end{tabular}
\end{table}

\begin{table}[htbp]
\centering
\small
\setlength{\tabcolsep}{3pt}
\caption{Details of LPO performance on AlignBench trained on infinity-instruct-1w dataset}
\label{tab:performance_alignbench_on_instruct}
\begin{tabular}{cccc}
\hline
Task & SFT & DPO & LPO \\ \hline
Professional Skill & \textbf{6.62} & 6.29 & 6.12 \\
Chinese Comprehension & \textbf{5.82} & 5.74 & 5.77 \\
Basic Task & \textbf{6.45} & 5.89 & 6.16 \\
Math Computation & 6.45 & 6.07 & \textbf{6.99} \\
Text Writing & 5.65 & 5.86 & \textbf{6.65} \\
Comprehensive Q\&A & 6.23 & \textbf{7.18} & 6.07 \\
Roleplay & \textbf{6.55} & 5.59 & 5.69 \\
Logical Reasoning & \textbf{5.66} & 5.14 & 5.38 \\
Chinese Reasoning & 6.06 & 5.61 & \textbf{6.18} \\
Chinese Language & \textbf{6.22} & 6.09 & 5.91 \\ \hline
\textbf{Overall Score} & \textbf{6.14} & 5.85 & 6.05 \\ \hline
\end{tabular}
\end{table}

Shown in Table \ref{tab:performance_mt_bench_on_perference}, \ref{tab:performance_alignbench_on_perference},  \ref{tab:performance_mt_bench_on_instruct} and \ref{tab:performance_alignbench_on_instruct}, the LPO algorithm achieves significant improvements over the SFT model: a 6.37\% gain on MT-bench and a 2.24\% improvement on Alignbench when trained on the Infinity-Preference dataset. Additionally, there is a 4.81\% enhancement on MT-bench when using the Infinity-Instruct-1w dataset.

While the model trained with DPO on the Infinity-Preference dataset showed notable gains, its performance on the Infinity-Instruct-1w dataset exhibited a slight decline on MT-bench and a more pronounced drop on Alignbench. Further analysis reveals that paired data constructed from Infinity-Preference exhibits finer distinctions and presents greater learning challenges. In contrast, pairs derived from Infinity-Instruct-1w display more pronounced differentiability and lower learning difficulty.

These findings indicate that the LPO algorithm demonstrates superior robustness, delivering consistent gains across diverse datasets. In contrast, DPO exhibits higher sensitivity to training data and appears prone to overfitting on simpler datasets. LPO demonstrates substantial gains in logical reasoning, while DPO shows more significant improvements in scenarios like Q\&A. We have further assessed the model’s effectiveness specifically within the mathematics vertical.

\subsection{Results on Math Tasks} \label{test_on_gsm8k_task_lpo}

We initialize our model using the SFT model that has been pre-trained on general tasks. During the alignment phase, we train the model using the dataset constructed via step-DPO \cite{lai2024step}. Detailed experimental configurations are provided in the Appendix \ref{lpo_ex_setting}.

We conducted extensive tests on the GSM8K \cite{cobbe2021gsm8k} benchmark. To better reflect real-world usage scenarios, we employed a zero-shot approach during inference. Here, “Qwen2.5-Instruct” denotes the officially released model from the Qwen team. The experimental results are presented in Table \ref{tab:lpo_performance_gsm8k}.

\begin{table}[ht]
\centering
\caption{Model performance on GSM8K}
\label{tab:lpo_performance_gsm8k} 
\begin{tabular}{cc}
\hline
Model Version & GSM8K \\ \hline
Qwen2.5-Instruct & 87.19 \\
SFT & 84.15 \\
DP0 & 82.34 \\
LP0 & \textbf{88.86} \\ \hline
\end{tabular}
\end{table}

As shown in Table \ref{tab:lpo_performance_gsm8k}, LPO achieves a score of 88.86 on the GSM8K benchmark, representing a 4.71-point improvement over the SFT model and surpassing the performance of Qwen2.5-Instruct. In contrast, DPO exhibits a 1.81-point degradation compared to the SFT baseline. As noted in DPOP, DPO often fails to achieve strong results on mathematical reasoning tasks.

\subsection{Results on Text-to-Speech Tasks}

In speech generation, synthesized audio sequences typically require excessively long token representations. We have validated LPO’s modeling capabilities under these long-sequence conditions. Building upon the Qwen-2.5-7B foundation, we expanded its audio token capacity and conducted incremental pre-training using 322B text and speech tokens. The model was further refined through instruction tuning on 440k TTS training samples (denoted as UniTTS-SFT), followed by specialized alignment-oriented LPO training (denoted as UniTTS-LPO). For details on incremental pre-training and instruction tuning, please refer to UniTTS \cite{wang2025unitts}.

The LPO training dataset is constructed as follows: Using each sample’s prompt as input, we generate three candidate responses. These candidates are then paired with the sample’s reference answer to form three preference pairs, which comprise the final training data. The corresponding training parameters are detailed in Appendix \ref{lpo_ex_setting}.

To evaluate the model's performance, it is assessed across the following dimensions, each rated on a scale of 0 to 5:

\begin{itemize}
    \item \textbf{Fidelity}: The degree to which the audio accurately reproduces the original sound, including the closeness of timbre, pitch, and other acoustic characteristics to the real voice.

    \item \textbf{Stability}: The absence of playback issues such as stuttering, frame skipping, or sudden interruptions
during audio playback.

    \item \textbf{Naturalness}: The quality of sounding like natural speech or playing, without noticeable robotic artifacts or an unnatural feel.

    \item \textbf{Emotional Expression}: The ability of the audio to accurately convey the intended emotions, such as joy, sadness, anger, etc.
\end{itemize}

\begin{table*}[htbp]
\centering
\caption{Comparison of UniTTS-SFT and UniTTS-LPO models}
\label{tab:tts_model_comparison}
\begin{tabular}{ccccc}
\hline
Model & Fidelity & Stability & Naturalness & Emotional expressiveness \\
\hline
UniTTS-SFT & 4.43 & \textbf{5} & 4.77 & 4.23 \\
UniTTS-LPO & \textbf{4.8} & 4.97 & \textbf{4.94} & \textbf{4.6} \\
\hline
\end{tabular}
\end{table*}

Table \ref{tab:tts_model_comparison} shows that the LPO algorithm demonstrates significant improvements in emotional expressiveness and fidelity compared to the SFT model, while exhibiting a slight decrease in stability. This outcome validates the effectiveness of the LPO algorithm in the field of speech generation.

\subsection{Results on Automatic Speech Recognition Task}

Automatic Speech Recognition (ASR) refers to the process of transforming spoken words into textual form, a task that involves detecting words within audio inputs and converting them into written language. The objective is to achieve a precise conversion of speech into text.

To assess the LPO’s performance on ASR, we conducted sequential SFT and LPO training on the expanded Qwen-2.5-7B model, leveraging both AISHELL-1 (Chinese) and LibriSpeech (English) corpora.

The LPO training data were prepared using two different methodologies:
\begin{itemize}
    \item \textbf{Model-based}: Inspired by TTS data synthesis methodologies, we leveraged the SFT model to generate LPO reject samples, which were then paired with reference samples.
    \item \textbf{Perturbation-based}: Text perturbation strategies \cite{Li_2022} were adopted to create reject samples through systematic corruption of reference sentences, employing insertion, deletion, and repetition operators at controlled noise ratios (\(\eta = 0.1\)).
\end{itemize}

Reject samples derived from these two construction methods exhibit divergent properties: the model-based method primarily generates homophonic heterographs (phonetically identical or similar but semantically distinct words), while the perturbation-based method produces samples with quantifiable stochasticity.

Note that for both model-based and perturbation-based methods, it is necessary to exclude data pairs where the chosen and reject samples are the same. Furthermore, data quality has a far greater impact on model performance than data quantity. In our experiments, we found that 1k high-quality samples contributed more significantly to the model's effectiveness and stability than 200k ordinary samples.

We use Character Error Rate (CER) for evaluating Chinese speech recognition and Word Error Rate (WER) for English.

Table \ref{tab:asr_model_comparison} shows that although constrained by the base model's fundamental capabilities, our model did not achieve state-of-the-art (SOTA) performance during SFT; however, the LPO algorithm effectively reduced the speech recognition error rate.

\begin{table*}[htbp]
\centering
\caption{Comparison of ASR-SFT and ASR-LPO models}
\label{tab:asr_model_comparison}
\begin{tabular}{|c|c|c|c|c|}
    \hline
    \multirow{2}{*}{Benchmark} & \multicolumn{3}{c|}{ASR-LPO} & ASR-SFT \\
    \cline{2-5} 
                       & Candidate method & LPO $r_2$ & CER/WER (\%) & CER/WER (\%) \\
    \hline
    \multirow{6}{*}{AISHELL-1} & \multirow{3}{*}{Model-based} & 1.0 & 3.583 & \multirow{6}{*}{3.868} \\
    \cline{3-4}
            &           & 2.0 & 3.621 & \\
    \cline{3-4}
            &           & 3.0 & 3.655 & \\
    \cline{2-4}
    & \multirow{3}{*}{Perturbation-based} & 1.0 & 3.583 & \\
    \cline{3-4}
            &           & 2.0 & \textbf{3.567} & \\
    \cline{3-4}
            &           & 3.0 & 3.694 & \\
    \hline
    \multirow{6}{*}{LibriSpeech-test-clean} & \multirow{3}{*}{Model-based} & 1.0 & 6.81 & \multirow{6}{*}{7.222} \\
    \cline{3-4}
            &           & 2.0 & 6.927 & \\
    \cline{3-4}
            &           & 3.0 & 6.927 & \\
    \cline{2-4}
    & \multirow{3}{*}{Perturbation-based} & 1.0 & 6.965 & \\
    \cline{3-4}
            &           & 2.0 & 6.874 & \\
    \cline{3-4}
            &           & 3.0 & \textbf{6.684} & \\
    \hline
\end{tabular}
\end{table*}

\subsection{Analysis of multi-epoch with different $r_2$}

\textbf{Analysis of Overfitting Phenomenon}: During DPO training, models are prone to overfitting and typically require both reduced learning rates and early stopping mechanisms. To verify whether the LPO algorithm exhibits similar susceptibility to overfitting, we replicated the experimental setup from the mathematics-specific chapter. This involved evaluating model performance on the GSM8K task across varying training epochs, while maintaining zero-shot inference during assessment.

\begin{figure}[htbp]
\centering
\includegraphics[scale=0.3]{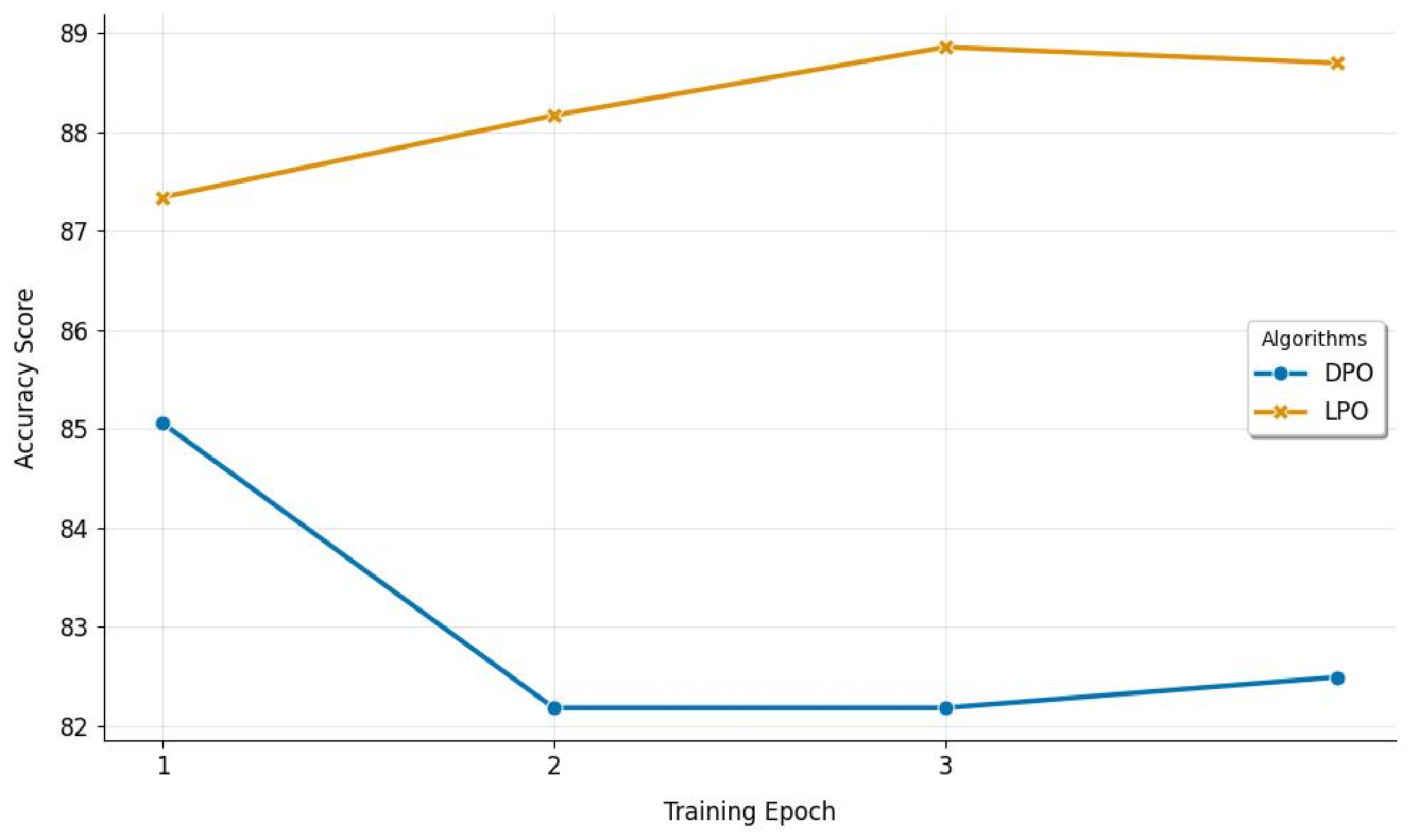}
\caption{GSM8K scores over training epochs on math tasks.}
\label{lpo_gsm8k_multi_epoch}
\end{figure}

As shown in Fig.\ref{lpo_gsm8k_multi_epoch}, DPO achieves its best performance in the first epoch but drops rapidly after the second epoch, even falling below the SFT model performance. In contrast, LPO shows steady improvement over the first three epochs, reaching its peak at the third epoch. This comparison demonstrates that LPO is less prone to overfitting compared to DPO.

\textbf{The influence of the coefficient of determination of \(r_2\)}:

In the algorithm analysis section, we demonstrate how the \(r_2\) coefficient regulates the rate of decline for rejected responses and the rate of increase for chosen responses, thereby modifying the model's performance. We validated the experimental outcomes for different \(r_2\) coefficients across both general tasks and the mathematics-specific domain. 

\begin{figure}[h]
    \centering
    \begin{subfigure}[b]{0.45\textwidth}
        \centering
        \includegraphics[width=\linewidth]{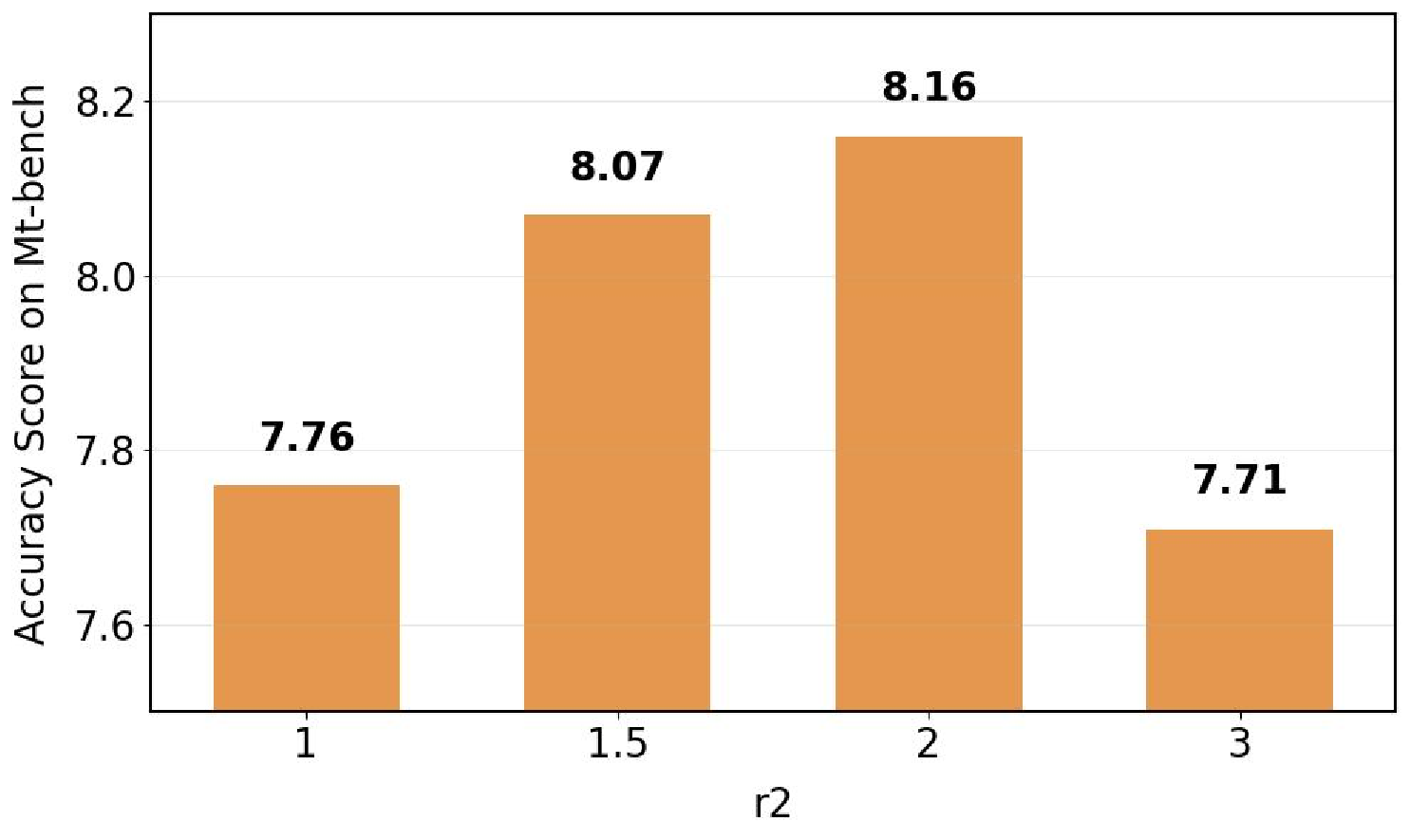}
        \caption{}
        \label{fig:r2_a}
    \end{subfigure}
    \hfill 
    \begin{subfigure}[b]{0.45\textwidth}
        \centering
        \includegraphics[width=\linewidth]{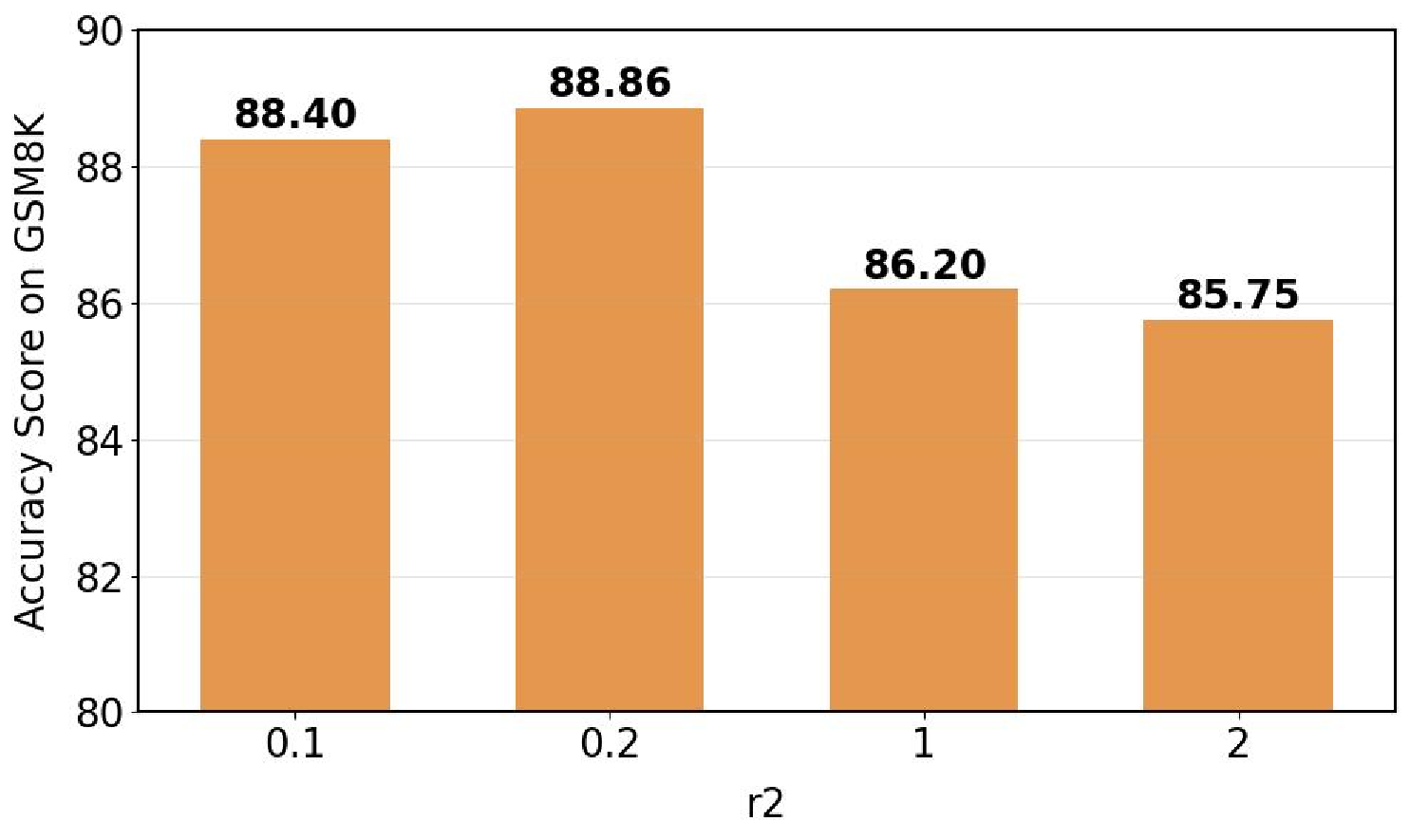}
        \caption{}
        \label{fig:r2_b}
    \end{subfigure}
    
    \caption{We tested the variation of model performance with the r2 coefficient: (a) Performance on the MT-Bench leaderboard for general tasks as r2 varies; (b) Performance on the GSM8K leaderboard for math tasks as r2 varies.}
    \label{fig:combined}
\end{figure}

For the general tasks, following the experimental setup detailed in Section \ref{test_on_general_task_lpo}, \(r_2\) coefficients were selected from the range [1, 1.5, 2, 3]. For vertically specific math tasks, using the configuration described in Section \ref{test_on_gsm8k_task_lpo}, \(r_2\) coefficients were tested at values of [0.1, 0.2, 1, 2]. The corresponding experimental results are illustrated in Fig. \ref{fig:r2_a} and Fig. \ref{fig:r2_b}.

As shown in Fig. \ref{fig:r2_a} and Fig. \ref{fig:r2_b}, our experimental results demonstrate two key conclusions: 1) Model performance varies on the leaderboard with different r2 coefficients. Thus, adjusting the r2 coefficient is necessary to prevent a rapid decline in rejection rate that causes overfitting. 2) The difficulty of learning varies across tasks. The r2 coefficient should be adjusted based on the rate of loss decrease for different tasks.

\section{Conclusion}

In this work, we first identify a critical limitation in DPO training: the simultaneous degradation of probabilities for both chosen and rejected responses during optimization. To address this issue, we propose the LPO algorithm, which decouples gradient control for chosen and rejected responses via the Straight-Through Estimator (STE). Our method regulates the rejection probability descent rate through parameter \(r_2\) while incorporating a positive reinforcement term to ensure monotonic improvement in the chosen response probability. Experimental results demonstrate consistent performance gains across general NLP tasks, specialized mathematical domains, and text-to-speech (TTS) applications, confirming LPO's robustness and broad applicability.

\bibliography{iclr2025_conference}
\bibliographystyle{iclr2025_conference}

\appendix
\section{Alignment Training Framework Development}

We conducted SFT and alignment training using the pai-megatron-patch framework. Since the framework lacks native support for alignment algorithms like DPO and LPO, we implemented custom modifications with the following key enhancements:

1) Added DPO/LPO algorithm support: The upgraded training pipeline now handles million-scale alignment datasets efficiently through mmap-format data loading, enabling rapid training initialization.

2) Extended multimodal capabilities: Beyond text modality, we implemented comprehensive speech modality support—including dataset construction, loading pipelines, and training workflows—with distributed inference during data preprocessing.

We've open-sourced this enhanced training framework to facilitate community adoption, enabling researchers to build upon our implementation or reproduce paper results.

\section{SFT Experimental Setup} \label{sft_configurations}

Infinity-Instruct is an open-source, high-quality dataset. We selected a subset of 290K training samples from it for supervised fine-tuning (SFT). The base model used was Qwen2.5-7B, with detailed training parameters provided in Table \ref{tab:sft_parameters}.
\begin{table}[ht]
\centering
\caption{Model training parameters for general task} 
\label{tab:sft_parameters} 
\begin{tabular}{cc}
\hline
Parameter Name & Parameter Value \\
\hline
BATCH\_SIZE & 128 \\
LR & 9e-6 \\
\hline
\end{tabular}
\end{table}

\section{LPO Experimental Setup} \label{lpo_ex_setting}

Table \ref{tab:param_lpo_general} presents the experimental settings for LPO across general tasks, domain-specific mathematical tasks, and TTS tasks.

\begin{table}[ht]
\centering
\caption{Training parameters for LPO}
\label{tab:param_lpo_general} 
\begin{tabular}{cccc}
\hline
Parameter Name & General Task & Math Task & TTS Task\\ \hline
R1 & 1.0 & 1.0 & 1.0 \\
R2 & 2.0 & 0.2 & 0.4 \\
BATCH\_SIZE & 24 & 24 & 120 \\
$\beta$ & 0.2 & 0.2 & 0.2\\
$\gamma$ & 10.0 & 10.0 & 10.0\\
LR & 2e-7 & 2e-7 & 2e-7 \\ \hline
\end{tabular}
\end{table}

\end{document}

%% file: math_commands.tex
\usepackage{amsmath,amsfonts,bm}

\def\eqref#1{equation~\ref{#1}}

\def\1{\bm{1}}

\DeclareMathAlphabet{\mathsfit}{\encodingdefault}{\sfdefault}{m}{sl}
\SetMathAlphabet{\mathsfit}{bold}{\encodingdefault}{\sfdefault}{bx}{n}

